\documentclass{article}
\usepackage{spconf,amsmath,amssymb,graphicx}
\usepackage{url}
\usepackage{xspace}
\usepackage{booktabs}
\usepackage{graphicx}
\usepackage{multirow}

\newcommand{\ie}{\emph{i.e.}\xspace}
\newcommand{\eg}{\emph{e.g.}\xspace}
\newcommand{\etal}{\emph{et al.}\xspace}

\title{Improved Regularization Techniques for End-to-End Speech Recognition}
\name{Yingbo Zhou, Caiming Xiong, Richard Socher}
\address{Salesforce Research}
\begin{document}
%
\maketitle
%


\begin{abstract}
Regularization is important for end-to-end speech models, since the models are highly flexible and easy to overfit. Data augmentation and dropout has been important for improving end-to-end models in other domains. However, they are relatively under explored for end-to-end speech models. Therefore, we investigate the effectiveness of both methods for end-to-end trainable, deep speech recognition models. 
We augment audio data through random perturbations of tempo, pitch, volume, temporal alignment, and adding random noise.
We further investigate the effect of dropout when applied to the inputs of all layers of the network. 
We show that the combination of data augmentation and dropout give a relative performance improvement on both Wall Street Journal (WSJ) and LibriSpeech dataset of over $20\%$. 
Our model performance is also competitive with other end-to-end speech models on both datasets.

\end{abstract}


\section{Introduction}
\label{sec:intro}
Regularization has proven crucial to improving the generalization performance of many machine learning models.
In particular, regularization is crucial when the model is highly flexible (\eg deep neural networks) and likely to overfit on the training data. Data augmentation is an efficient and effective way of doing regularization that introduces very small (or no) overhead during training. It has shown to consistently improve performance in various pattern recognition tasks  \cite{alex_imagenet,ko2015audio,Cui2014feataug,Cui2015feataug,jaitly2013vocal}. 
Dropout \cite{dropout} is another powerful way of doing regularization for training deep neural networks, 
it intends to reduce the co-adaptation amongst hidden units by randomly zero-ing out inputs to the hidden layer during training.

End-to-end speech models often have millions of parameters \cite{miao2015eesen,hannun2014deep,amodei2016deep,bahdanau2016end,zhang2017very}. 
However, data augmentation and dropout have not been extensively studied or applied to them. 
We investigate the effectiveness of data augmentation and dropout for regularizing end-to-end speech models. 
In particular, we augment the raw audio data by changing the tempo and pitch independently.
The volume and temporal alignment of the audio signals are also randomly perturbed, with additional random noises added.
To further regularize the model, we employ dropout to each input layer of the network. 
With these regularization techniques, we obtained over $20\%$ relative performance on the Wall Street Journal (WSJ) dataset and LibriSpeech dataset.

\section{Related Work}
Data augmentation in speech recognition has been applied before. 
Gales \etal \cite{gales2009support} use hidden Markov models to generate synthetic data to enable 1-vs-1 training of SVMs. 
Feature level augmentation has also demonstrated effectiveness \cite{jaitly2013vocal,Cui2014feataug,Cui2015feataug}. 
Ko \etal \cite{ko2015audio} performed audio level speed perturbation that also lead to performance improvements.
Background noise is used for augmentation in \cite{hannun2014deep,amodei2016deep} to improve performance on noisy speech. Apart from adding noise, our data augmentation also modifies the tempo, pitch, volume and temporal alignment of the audio.
Dropout has been applied to speech recognition before. It has been applied to acoustic models in \cite{dahl2013improving,sainath2015deep,saon2015ibm} and demonstrated promising performance. 
For end-to-end speech models, Hannun \etal \cite{hannun2014deep} applied dropout to the output layer of the network.
However, to the best of our knowledge, our work is the first to apply it to other layers in end-to-end speech models.

\section{Model Architecture}
\label{sec:model_sturcture}
\begin{figure}[tbp!]
\centering
\includegraphics[width=0.8\columnwidth,trim=2.1in 1.2in 4.8in 0.4in,clip]{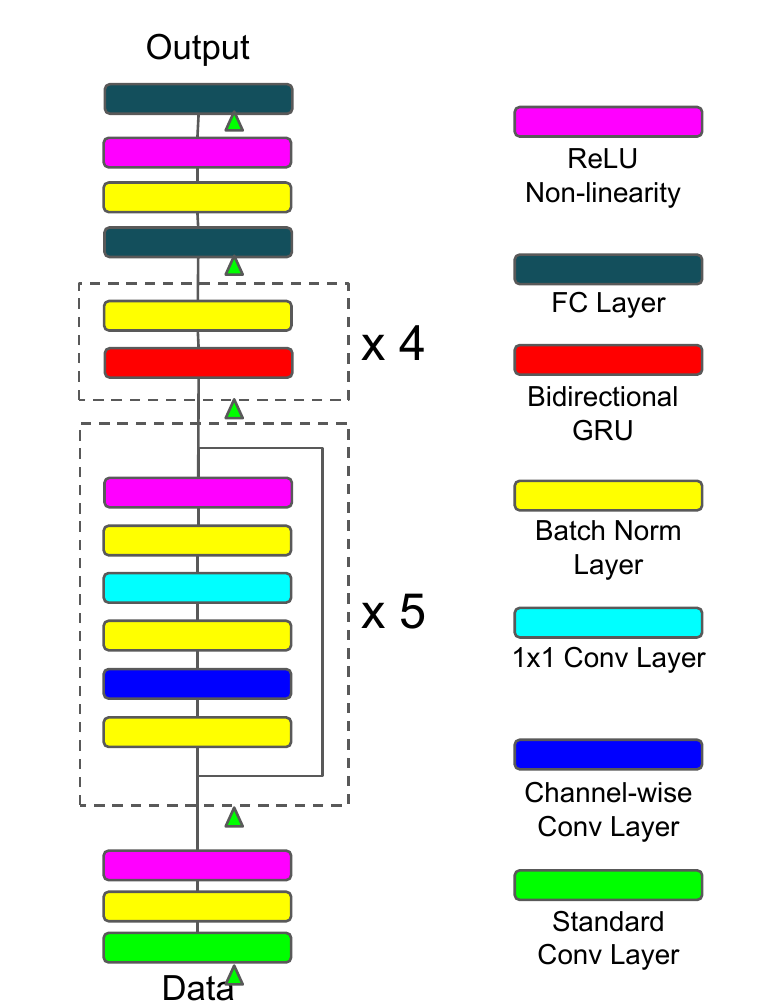}
\caption{Model architecture of our end-to-end speech model. Different colored blocks represent different layers as shown on the right, the triangle indicates dropout happens right before the pointed layer.}
\label{fig:model_structure}
\end{figure}

The end-to-end model structure used in this work is very similar to the model architecture of Deep Speech 2 (DS2) 
\cite{amodei2016deep}. While we have made some modifications at the front-end time-frequency convolution (\ie 2-D convolution) layers,
the core structure of recurrent layers is the same. 
The full end-to-end model structure is illustrated in Fig.~\ref{fig:model_structure}. 

First, we use depth-wise separable convolution \cite{sifre2013rotation, chollet2016xception} for all the convolution layers. 
The performance advantage of depth-wise separable convolution has been demonstrated in computer vision tasks \cite{chollet2016xception} and is also more computationally efficient. 
The depth-wise separable convolution is implemented by first convolving over the input channel-wise. It then convolves with $1\times1$ filters with the desired number of channels. 
Stride size only influence the channel-wise convolution; the following $1\times 1$ convolutions always have stride one. Second, we substitute normal convolution layers with ResNet \cite{he2016deep} blocks. The residual connections help the gradient flow during training. They have been employed in speech recognition \cite{zhang2017very} and achieved promising results. For example, a $w\times h$ depth-wise separable convolution with $n$ input and $m$ output channels is implemented by first convolving the input channel-wise with its corresponding $w\times h$ filters, followed by standard $1\times 1$ convolution with $m$ filters. 

%

Our model is composed of one standard convolution layer that has larger filter size, followed by five residual convolution blocks. 
Convolutional features are then given as input to a 4-layer bidirectional recurrent neural network with gated recurrent units (GRU) \cite{cho2014properties}. 
Finally, two fully connected layers take the last hidden RNN layer as input and output the final per-character prediction.
Batch normalization \cite{bn} is applied to all layers to facilitate training. 

\section{Preventing Overfitting}
Our model has over five million parameters (see sec. \ref{sec:exp}). This makes regularization important for it to generalize well.
In this section we describe our primary methods for preventing overfitting: data augmentation and dropout.

\subsection{Data Augmentation}
Vocal tract length perturbation (VTLP \cite{jaitly2013vocal}) is a popular method for doing feature level data augmentation in speech. We choose to do data level augmentation (\ie augment raw audio) instead of feature level augmentation, because the absence of feature-level dependencies makes it more flexible. 
Ko \etal \cite{ko2015audio} used data level augmentation and showed that modifying the speed of raw audio approximates the effect of VTLP and works better than VTLP. 
However, in speed perturbation since the pitch is positively correlated with the speed, it is not possible to generate audio with higher pitch but slower speed and vice versa. 
This may not be ideal, since it reduces the variation in the augmented data, which
in turn may hurt performance. 
Therefore, To get increased variation in our data,
we separate the speed perturbation into two independent 
components 
-- tempo and pitch. 
By keeping the pitch and tempo separate, we 
can cover a wider range of variations. We use 
the \texttt{tempo} and \texttt{pitch} functions
from the SoX audio manipulation tool \cite{sox}.  

Generating noisy versions
of the data is also a common way to do data augmentation. To generate such data, we add random white noise to the audio signal. Volume of the audio is also randomly modified to simulate the effect of different recording volumes. To further distort the audio, it is also randomly shifted by a small amount (\ie less than 10ms). With 
a combination of the above approaches, we can synthetically generate a large amount of data that captures different variations.


\subsection{Dropout}
Dropout \cite{dropout} is a powerful regularizer. It prevents the co-adaptation of hidden units by randomly zero-ing out a subset of inputs for that layer during training. In more detail, let $\mathbf{x}_i^t \in \mathbb{R}^{d}$ denote the $i$-th input sample to a network layer at time $t$, dropout does the following to the input during training
\begin{align}
z_{ij}^t &\sim \text{Bernoulli}(1-p) \qquad j \in \{1,2,\ldots,d\}\\
\hat{\mathbf{x}}_i^t &= \mathbf{x}_i^t \odot \mathbf{z}_i^t
\end{align}
where $p$ is the dropout probability, $\mathbf{z}_i^t = [z_{i1}^t, z_{i2}^t, \ldots, z_{id}^t]$ is the dropout mask for $\mathbf{x}_i^t$, and $\odot$ denote elementwise multiplication. At test time, the input is rescaled by $1-p$ so that the expected pre-activation stays the same as it was at training time. This setup works well for feed forward networks in practice, however, it hardly finds any success when applied to recurrent neural networks.

Gal and Ghahramani \cite{gal2016theoretically} proposed a dropout variant that approximates a Bayesian neural network for recurrent networks. 
The modification is principled and simple, \ie instead of randomly drop different dimensions of the input across time, a fixed random mask is used for the input across time. More precisely, we modify the dropout to the input as follows:\footnote{In \cite{gal2016theoretically}, the authors also have dropout applied on recurrent connections, we did not employ the recurrent dropout because dropout on the input is an easy drop-in substitution for cuDNN RNN implementation, whereas the recurrent one is not. Switch from cuDNN based RNN implementation will increase the computation time of RNNs by a significant amount, and thus we choose to avoid it.}
\begin{align}
\label{eq:drop1}
z_{ij} &\sim \text{Bernoulli}(1-p) \qquad j \in \{1,2,\ldots,d\}\\
\label{eq:drop2}
\hat{\mathbf{x}}_i^t &= \mathbf{x}_i^t \odot \mathbf{z}_i 
\end{align}
where $\mathbf{z}_i = [z_{i1}, z_{i2}, \ldots, z_{id}]$ is the dropout mask. Since we are not interested in the Bayesian view of the model, we choose the same rescaling approximation as standard dropout (\ie rescale input by $1-p$) instead of doing Monte Carlo approximation at test time.

We apply the dropout variant described in eq. \ref{eq:drop1} and \ref{eq:drop2} to inputs of all convolutional and recurrent layers.
Standard dropout is applied on the fully connected layers.

\section{Experiments}
\label{sec:exp}
To investigate the effectiveness of the proposed techniques, we perform experiments on the Wall Street Journal (WSJ) and LibriSpeech \cite{panayotov2015librispeech} datasets. The input to the model is a spectrogram computed with a 20ms window and 10ms step size.
We normalize each spectrogram to have zero mean and unit variance. In addition, we also normalize each feature to have zero mean and unit variance based on the training set statistics. No further preprocessing is done after these two steps of normalization. 

We denote the size of the convolution layer by tuple $(\text{C, F, T, SF, ST})$, where C, F, T, SF, and ST denotes number of channels, filter size in frequency dimension, filter size in time dimension, stride in frequency dimension and stride in time dimension respectively. We have one convolutional layer with size (32,41,11,2,2), and five residual convolution blocks of size (32,7,3,1,1), (32,5,3,1,1), (32,3,3,1,1), (64,3,3,2,1), (64,3,3,1,1) respectively. Following
the convolutional layers we have 4-layers of bidirectional GRU RNNs with $1024$ hidden units per direction per layer. Finally we have one fully connected hidden layer 
of size $1024$ followed by the output layer. The convolutional and fully connected layers are initialized uniformly \cite{he2015delving}. The recurrent layer weights are initialized with a uniform distribution $\mathcal{U}(-1/32, 1/32)$. The model is trained in an end-to-end fashion to maximize the log-likelihood
using connectionist temporal classification \cite{graves2006ctc}. We use mini-batch stochastic gradient descent with batch size 64, learning rate 0.1, and with Nesterov momentum 0.95. The learning rate is reduced by half whenever the validation loss has 
plateaued, 
and the model is trained until the validation loss stops improving. The norm of the gradient is clipped \cite{pascanu2013difficulty} to have a maximum value of $1$.
In addition, for all experiments we use $l$-2 weight decay of $1e^{-5}$ for all parameters.

\subsection{Effect of individual regularizer}
To study the effectiveness of data augmentation and dropout, we perform experiments on both datasets with various settings. The first set of experiments were carried out on the WSJ corpus. We use the standard \emph{si284} set for training, \emph{dev93} for validation and \emph{eval92} for test evaluation. We use the provided language model and report the result in the 20K closed vocabulary setting with beam search. 
The beam width is set to 100. Since the training set is relatively small ($\sim80$ hours), we performed
a more detailed ablation study on this dataset by separating the tempo based augmentation from
the one that generates noisy versions of the data. For tempo based data augmentation, the tempo parameter is selected
following a uniform distribution $\mathcal{U}(0.7, 1.3)$, and $\mathcal{U}(-500,500)$ for pitch. Since WSJ has relatively clean recordings, we keep the signal to noise ratio between 10 and 15db when adding white noise. The gain is selected from $\mathcal{U}(-20, 10)$ and the audio is shifted randomly by 0 to 10ms. Results are shown in Table~\ref{tbl:wsj_aug_comp}. Both approaches improve
the performance over the baseline, where none of the additional regularization is applied. Noise augmentation has demonstrated its effectiveness for  making the model more robust against noisy inputs. We show here that adding a small amount of noise also benefits the model on relatively clean speech samples. To compare with existing augmentation methods,
 we trained a model by using speed perturbation.
We use 0.9, 1.0, and 1.1 as the perturb coefficient for speed as suggested in \cite{ko2015audio}. This results in
a WER of 7.21\%, which brings 13.96\% relative performance improvement. Our tempo based augmentation is slightly better than the speed augmentation, which may attribute to more variations in the augmented data. When the techniques for data augmentation are combined, we have a significant relative improvement of $20\%$ over the baseline (see table \ref{tbl:wsj_aug_comp}). 
\begin{table}[t!]
\centering
\begin{tabular}{lcc}
\toprule
Regularization Methods & \emph{eval92} & Rel. Improvement \\
\midrule
Baseline & 8.38\% & --\\
$+$Noise & 7.88\% & 5.96\%\\
$+$Tempo augmentation & 7.02\% & 16.22\%\\
$+$All augmentation & 6.63\% & 20.88\%\\
$+$Dropout & 6.50\% & 22.43\%\\
$+$All regularization & 6.42\%& 23.39\%\\
\bottomrule
\end{tabular}
\caption{Word error rate from WSJ dataset. Baseline denotes model trained only with weight decay; noise denotes model trained with noise augmented data; tempo augmentation denotes model trained with independent tempo and pitch perturbation; all augmentation denotes model trained with all proposed data augmentations; dropout denotes model trained with dropout.}
\label{tbl:wsj_aug_comp}
\end{table}

\begin{figure}[tbp!]
\centering
\begin{tabular}{cc}
\includegraphics[width=0.5\columnwidth]{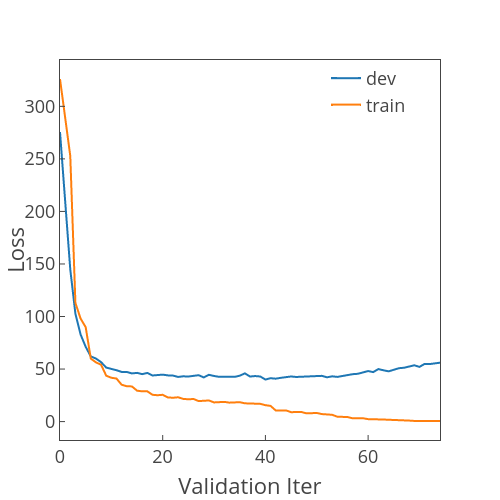} &
\includegraphics[width=0.5\columnwidth]{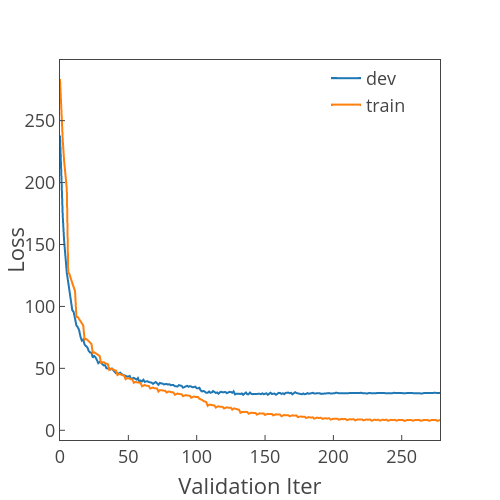}\\
(a) & (b)\\
\end{tabular}
\caption{Training and validation loss on WSJ dataset, where a) shows the learning curve from the baseline model, and b) shows the loss when regularizations are applied.}
\label{fig:training_curve}
\end{figure}

Dropout also significantly improved the performance ($22\%$ relative improvement, see table \ref{tbl:wsj_aug_comp}). The dropout probabilities are set as follows, $0.1$ for data, $0.2$ for all convolution layers, $0.3$ for all recurrent and fully connected layers.
By combining all regularization, we achieve a final WER of $6.42\%$. Fig. \ref{fig:training_curve} shows the training curve of baseline and regularized models. It is clear that with regularization, the gap between the validation and training loss is narrowed. In addition, the regularized training also results in a lower validation loss.

We also performed experiments on the LibriSpeech dataset. The model is trained using all 960 hours of training data.
We use both \emph{dev-clean} and \emph{dev-other} for validation and report results on \emph{test-clean} and \emph{test-other}. The provided 4-gram language model is used for final beam search decoding.
The beam width used in this experiment is also set to 100. The results follow a similar trend as the previous experiments. We achieved relative performance improvement of over $22\%$ on test-clean and over $32\%$ on test-other set (see table \ref{tbl:libri}).

\begin{table}[t!]
\centering
\begin{tabular}{lcc}
\toprule
Regularization & \multirow{2}{*}{test-clean} & \multirow{2}{*}{test-other} \\
Methods & &\\
\midrule
Baseline & 7.45\% & 22.59\%\\
$+$All augmentation & 6.31\% (15.30\%) & 18.59\% (17.70\%)\\
$+$Dropout & 5.87\% (21.20\%) & 17.08\% (24.39\%)\\
$+$All regularization & 5.67\% (23.89\%) & 15.18\% (32.80\%)\\
\bottomrule
\end{tabular}
\caption{Word error rate on the Librispeech dataset, numbers in parenthesis indicate relative performance improvement over baseline. Notations are the same as in table \ref{tbl:wsj_aug_comp}.}
\label{tbl:libri}
\end{table}

\subsection{Comparison to other methods}
In this section, we compare our end-to-end model with other end-to-end speech models. The results from WSJ and LibriSpeech (see table \ref{tbl:wsj_comp} and \ref{tbl:libri_comp}) are obtained through beam search decoding with the language model provided with the dataset with beam size 100.
To make a fair comparison on the WSJ corpus, we additionally trained an extended trigram model with the data released with the corpus. 
Our results on both WSJ and LibriSpeech are competitive to existing methods. We would like to note that our model achieved comparable results with Amodei \etal \cite{amodei2016deep} on LibriSpeech dataset, although our model is only trained only on the provided training set.
This demonstrates the effectiveness of the proposed regularization methods for training end-to-end speech models.

\begin{table}[t!]
\centering
\begin{tabular}{lc}
\toprule
Method & Eval 92 \\
\midrule
Bahdanau \etal \cite{bahdanau2016end} & 9.30\% \\
Graves and Jaitly \cite{graves2014towards} & 8.20\% \\
Miao \etal \cite{miao2015eesen} & 7.34\% \\
Ours & 6.42\% \\
Ours (extended 3-gram)& 6.26\% \\
Amodei \etal \cite{amodei2016deep} & 3.60\%\\
\bottomrule
\end{tabular}
\caption{Word error rate comparison with other end-to-end methods on WSJ dataset.}
\label{tbl:wsj_comp}
\end{table}

\begin{table}[t!]
\centering
\begin{tabular}{lcc}
\toprule
Method & test-clean & test-other \\
\midrule
Collobert \etal \cite{collobert2016wav2letter} & 7.20\% & - \\
ours & 5.67\% & 15.18\% \\
Amodei \etal \cite{amodei2016deep} & 5.33\% & 13.25\% \\
\bottomrule
\end{tabular}
\caption{Word error rate comparison with other end-to-end methods on LibriSpeech dataset.}
\label{tbl:libri_comp}
\end{table}

\section{Conclusion}
In this paper, we investigate the effectiveness of data augmentation and dropout for deep neural network based, end-to-end speech recognition models.
For data augmentation, we independently vary the tempo and pitch of the audio so that it is able to generate a large variety of additional data. In addition, we also add noisy versions of the data by changing the gain, shifting the audio, and add random white noise. We show that, with tempo and noise based augmentation, we are able to achieve 15--20\% relative performance improvement on WSJ and LibriSpeech dataset.
We further investigate the regularization of dropout by applying it to inputs of all layers of the network. Similar to data augmentation, we obtained significant performance improvements. When both regularization techniques are combined, we achieved new state-of-the-art results on both dataset, with 6.26\% on WSJ, and 5.67\% and 15.18\% on test-clean and test-other set from LibriSpeech.

\bibliographystyle{IEEEbib}
\bibliography{refs}

\end{document}